\documentclass[sigconf]{acmart}

\AtBeginDocument{%
  }

\setcopyright{rightsretained}
\copyrightyear{2024}
\acmYear{2024}
\acmDOI{10.1145/3688868.3689199}

\acmConference[MCHM '24]{Proceedings of the 1st International Workshop on Multimedia Computing for Health and Medicine}{October 28-November 1, 2024}{Melbourne, VIC, Australia}
\acmBooktitle{Proceedings of the 1st International Workshop on Multimedia Computing for Health and Medicine (MCHM '24), October 28-November 1, 2024, Melbourne, VIC, Australia}
\acmISBN{979-8-4007-1195-4/24/10}

\begin{document}

\title{Real-Time Posture Monitoring and Risk Assessment for Manual Lifting Tasks Using MediaPipe and LSTM}


\author{Ereena Bagga}
\affiliation{%
  \institution{Deakin University}
  \city{Burwood}
  \country{Australia}}
\email{ereenab70@gmail.com}

\author{Ang Yang}
\affiliation{%
  \institution{Deakin University}
  \city{Waurn Ponds}
  \country{Australia}}
\email{leon.yang@deakin.edu.au}


\begin{abstract}
  This research focuses on developing a real-time posture monitoring and risk assessment system for manual lifting tasks using advanced AI and computer vision technologies. Musculoskeletal disorders (MSDs) are a significant concern for workers involved in manual lifting, and traditional methods for posture correction are often inadequate due to delayed feedback and lack of personalized assessment. Our proposed solution integrates AI-driven posture detection, detailed keypoint analysis, risk level determination, and real-time feedback delivered through a user-friendly web interface. The system aims to improve posture, reduce the risk of MSDs, and enhance user engagement. The research involves comprehensive data collection, model training, and iterative development to ensure high accuracy and user satisfaction. The solution's effectiveness is evaluated against existing methodologies, demonstrating significant improvements in real-time feedback and risk assessment. This study contributes to the field by offering a novel approach to posture correction that addresses existing gaps and provides practical, immediate benefits to users.
\end{abstract}



\maketitle

\section{Introduction}

Musculoskeletal disorders (MSDs) are highly prevalent in Australia, impacting 6.9 million people in 2014-15 \cite{oakman2019work}. In 2011, MSDs accounted for 12\% of the nation's total burden of disease and injury and 23\% of the non-fatal burden, ranking second only to mental health and substance use disorders. These disorders also have significant economic impacts on both society and individuals. Back pain and related problems constitute 31\% of Australia's MSD burden, with 17\% attributed to occupational exposures. Previous research has found that workplace hazards are responsible for 37\% of back pain cases globally. The economic burden of work-related musculoskeletal disorders (WMSDs) is considerable, encompassing health care costs, compensation, lost income, and early retirement. The Global Burden of Disease (GBD) Study in 2012 highlighted low back pain as the leading cause of physical disability worldwide, responsible for 83 million years lived with disability \cite{oakman2019work}. A more recent report shows that the situation has not improved for WMSDs worldwide \cite{chan2022role}. Consequently, primary WMSD prevention remains a key focus for researchers, practitioners, and organizations.

Advances in wearable sensing technologies and machine learning (ML) are proving particularly useful in addressing emerging research questions within work-related musculoskeletal disorder (WMSD) prevention themes \cite{yan2021applying} \cite{hussain2019digital} \cite{chen2000new}. ML, a key branch of artificial intelligence, is increasingly utilized across various industries due to technological advancements that enhance data collection and processing capabilities. ML involves using algorithms to optimize performance criteria based on training data and past experiences \cite{jordan2015machine} \cite{mahadevkar2022review} \cite{yang2015enhanced}. Researchers specify hyperparameters \cite{yang2018efficient} that guide ML algorithms to "learn" from existing data, developing models that reveal underlying structures (unsupervised learning) or make predictions about new data (supervised learning) \cite{nasteski2017overview} \cite{liu2024iot}. ML algorithms are versatile, and applicable for regression, classification, clustering, and reinforcement learning. Unsupervised learning can identify patterns in datasets, such as using k-means clustering to find subgroups. Reinforcement learning determines optimal action sequences to maximize returns and has been used to provide tailored feedback in virtual posture training environments for workers. Supervised learning, the most common approach, classifies or makes estimations on new cases, exemplified by the k-nearest neighbours (k-NN) classifier used for movement recognition. ML can model complex, non-linear interactions between numerous variables, making it well-suited for understanding the multifaceted etiology of WMSDs and aiding in their prevention. Despite a trade-off between interpretability and predictive performance, ML techniques are advancing primary WMSD prevention efforts \cite{wang2021computer} \cite{wang2023posture} \cite{jung2022computer} \cite{antwi2022deep} \cite{zhao2021applying}.

However, most existing methods often fail to provide timely and personalized feedback, limiting their effectiveness in preventing WMSDs \cite{fisher2023occupational}. Recent advances in pre-trained AI models for human posture estimation present new opportunities for developing more effective solutions for real-time posture monitoring and correction \cite{mukhaimar2023multi}. For instance, Google's MediaPipe \cite{lugaresi2019mediapipe} enables the detection of human body landmarks in images or videos, allowing for the identification of key body locations, posture analysis, and movement categorization using machine learning models that work with both single images and videos. This task outputs body pose landmarks in image coordinates and three-dimensional world coordinates.

This research aims to develop a comprehensive system leveraging MediaPipe and Long Short-Term Memory (LSTM) networks \cite{yu2019review} to monitor lifting postures in real-time, assess associated risks, and provide immediate corrective feedback. The system performs detailed keypoint analysis to categorize postures based on correct and incorrect movements. The study follows a multi-phase approach, including extensive data collection, model training, and iterative development. The research highlights its unique contributions and effectiveness by comparing the proposed solution with existing literature, setting the stage for a detailed exploration of the system's design, implementation, and evaluation.

Our main contributions are as follows: 1. We proposed an LSTM model for risky lifting posture detection. 2. We implemented a dataset consisting of videos of good and bad lifting actions at different angles and distances. 3. We implemented a system that allows to use of MediaPipe to estimate pose and use our proposed LSTM model to prevent WMSDs through a webcam. 

This paper is divided into the following sections: Section 2 reviews related work on preventing MSDs and pose estimation. Section 3 discusses relevant concepts and theories, providing more detail on our key contributions. Section 4 explains the procedures performed and experimental settings. Finally, Section 5 presents the conclusion of this research and offers recommendations for future work.

\section{Literature Review}

\subsection{Literature Synthesis}
The literature review underscores the gravity of musculoskeletal disorders (MSDs) resulting from manual lifting activities, providing a foundational understanding of the prevalent issues. Identified papers strongly support the significance of posture analysis, particularly through the Ovako Working Posture Analysis System (OWAS method), shedding light on the risks faced by workers in various industries (e.g. the construction industry)\cite{abraham2022virtual} \cite{marquez2019prediction}. Moreover, a shift towards technological solutions is evident, with research emphasizing the integration of advanced technologies, including computer vision, image processing, and machine learning, to address occupational health and safety (OHS) concerns. The synthesis reveals a convergence of efforts toward leveraging machine learning algorithms for posture estimation and recognition, highlighting the potential of adaptive systems to enhance accuracy. The pivotal role of computer vision and image processing techniques emerges as a core theme, focusing on real-time monitoring and analysis of postures during manual lifting.

\subsection{Identified Gaps}
The research identifies several critical gaps in posture recognition for manual lifting. Firstly, there's a lack of cohesive integration between machine learning and computer vision approaches, with few studies addressing their synergies explicitly \cite{fisher2023occupational}. Secondly, the literature primarily focuses on generic applications, neglecting in-depth analyses of specific industry contexts that are crucial for optimizing system effectiveness. Thirdly, there's a gap in understanding the long-term efficacy of technology-driven solutions, with limited studies evaluating sustained effectiveness in preventing Musculoskeletal Disorders (MSDs) over extended periods. Furthermore, the reliance on limited and controlled datasets highlights the need for more extensive data collection efforts to enhance model robustness and generalizability \cite{sun2022retracted}. Lastly, current models struggle with environmental and user variability, emphasizing the importance of developing adaptive algorithms for maintaining accuracy and effectiveness in real-world applications, particularly in dynamic manual lifting scenarios.

Existing solutions, such as manual monitoring systems and generic computer vision applications, have limitations that hinder their effectiveness in delivering instantaneous feedback and intervention. Human observers are prone to subjectivity and cannot consistently ensure timely correction, while basic computer vision models may lack adaptability and struggle to provide real-time insights. Therefore, there is a critical need for a more dynamic and responsive solution to address these shortcomings \cite{zhao2021applying}. Traditional manual training programs exhibit limited effectiveness as they often rely on periodic sessions, which may not maintain awareness and proper practices over time. Additionally, conducting regular, in-person training sessions is resource-intensive in terms of time and personnel. Generic ergonomic guidelines fall short of addressing the specific needs of diverse job roles or industries, leading to suboptimal outcomes. These guidelines also lack personalization, failing to account for individual variations in physical abilities or health conditions. The use of wearable sensors faces challenges such as limited acceptance among employees due to discomfort, privacy concerns, or perceptions of surveillance. Moreover, the effectiveness of wearable solutions heavily depends on the accuracy and reliability of the sensors used. Manual monitoring systems introduce subjectivity and may not provide objective and consistent assessments. Additionally, continuous manual monitoring requires dedicated personnel, making it resource-intensive and potentially impractical in large-scale operations.

Generic computer vision solutions may struggle to adapt to diverse environmental conditions, affecting accuracy. Implementing tailored computer vision systems can also incur high initial development costs, making it less accessible for smaller businesses \cite{antwi2022deep}. Existing solutions often lack real-time feedback mechanisms, resulting in delayed intervention and increased risk of musculoskeletal disorders. Without instant feedback, employees may continue unsafe practices without immediate correction. Web-based training modules without real-time monitoring lack the ability to actively engage employees during manual lifting activities. This limitation may result in inconsistent application of learned principles without continuous monitoring and reinforcement \cite{ciccarelli2023spectre}. Some existing solutions underutilize data analytics, which hinders the ability to identify trends, patterns, or areas for improvement in occupational safety. The absence of predictive analytics further limits the ability to proactively address potential risks before they escalate \cite{mahadevkar2022review}. Understanding these limitations is crucial for developing an intelligent solution that overcomes these challenges and provides a more effective and adaptive approach to occupational injury risk mitigation.

\section{System Design and Methodology}

\subsection{System Architecture}
The proposed system architecture integrates MediaPipe and an LSTM-based classifier to monitor and assess employees' postures during manual lifting. Real-time data capture via webcams enables the computer vision system to continuously process posture information. The LSTM model, trained on a diverse dataset encompassing various lifting techniques and postures, evaluates this data to identify correct or incorrect postures. The seamless integration of these components within the system architecture forms the backbone of an effective solution for preventing MSDs and promoting workplace safety. The workflow of the proposed model is shown in Figure 1.

\begin{figure}[h]
\centering
\includegraphics[width=\linewidth]{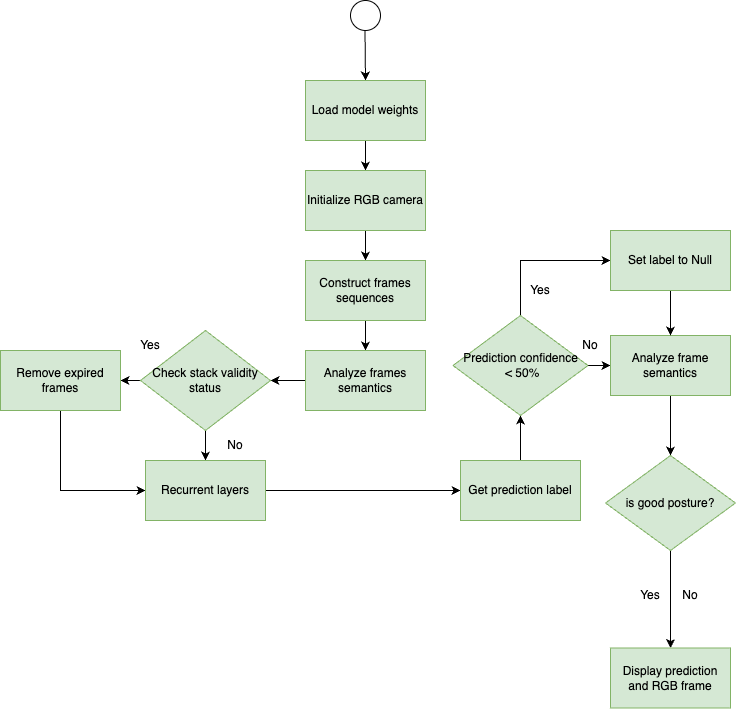}
\caption{Workflow diagram for a lifting posture state of a worker, depicting the flow of information from the webcam to the AI model.}
\end{figure}

The selection of this integrated solution is underpinned by several key considerations. Firstly, the accuracy provided by computer vision, particularly through the integration of MediaPipe, enables precise posture recognition. This accuracy is paramount for identifying proper postures during manual lifting and minimizing the risk of MSDs. Furthermore, the real-time monitoring capability of the system allows for immediate corrective actions in response to unsafe postures. This real-time aspect significantly enhances the effectiveness of the solution in preventing occupational injuries. The incorporation of a web component adds an interactive dimension to the solution. By displaying real-time posture results in an easily interpretable format, the web interface fosters user engagement and promotes continuous improvement in posture awareness and correction. Moreover, the system's scalability is ensured by its architecture, which leverages standard webcams for deployment across various work environments. The potential for future expansion to higher-resolution cameras and advanced computing resources underscores the system's adaptability to evolving needs.

LSTMs are well-suited for tasks involving sequential or time-series data \cite{phat2022proposing}, making them ideal for capturing temporal dependencies inherent in posture recognition during manual lifting activities. Additionally, LSTMs possess memory retention capabilities, allowing the model to consider past poses when predicting the current one. This contextual understanding is crucial for assessing the dynamics of movements and ensuring accurate posture classification. Furthermore, LSTMs excel in learning complex patterns within sequential data. Given the intricate variations in body movements during manual lifting, the LSTM model's ability to discern and analyze these patterns contributes significantly to the system's overall accuracy and effectiveness.

\subsection{Dataset Details}
The dataset used for this project consists of 62 videos, evenly divided into good and bad postures. These videos were collected from different fields of view, angles, and participants of varying heights. The model extracts keypoint values using the MediaPipe model and saves these keypoints in structured folders. This dataset encompasses various actions and sequences, providing a diverse training set. The detailed breakdown of the dataset is shown in Table 1.

\begin{table*}
\centering
\caption{Detailed Breakdown of Dataset.}
\begin{tabular}{ccccccc} 
    \toprule
    Posture Type & Number of Videos & Frames per Video & Landmarks per Frame & Total Keypoint Values per Frame \\
    \midrule
    Good Postures & 31 & 30 & 33 & 132 \\
    Bad Postures & 31 & 30 & 33 & 132  \\
    \midrule
\end{tabular}
\end{table*}


\begin{figure}[h]
\centering
\includegraphics[width=\linewidth]{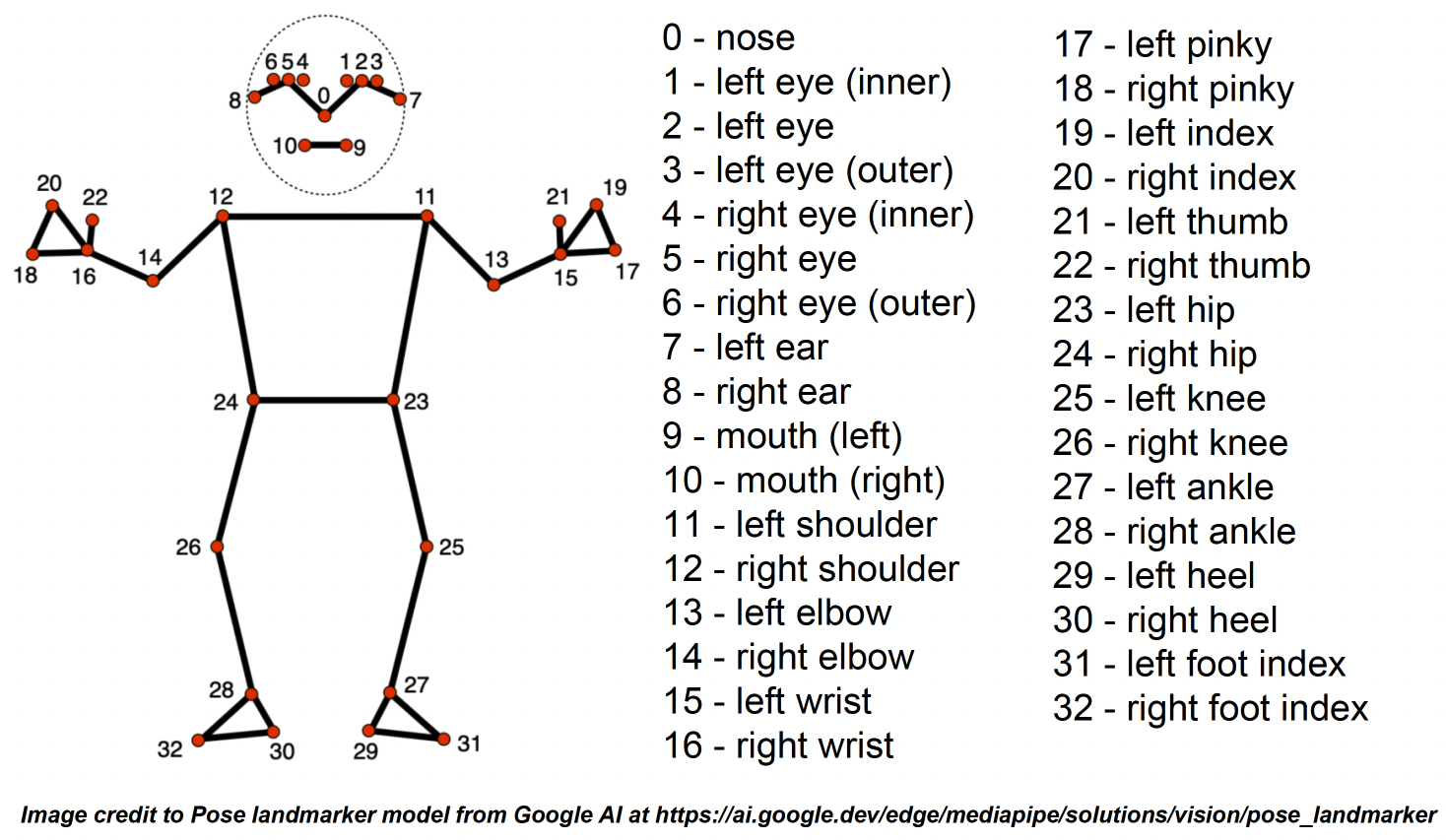}
\caption{Illustrates the identification of 33 landmarks on the human body through the utilization of Mediapipe.}
\end{figure}

\subsection{Implementation}

The initial part of the implementation involves real-time pose estimation using the MediaPipe Holistic model. This is achieved by capturing video frames, processing them through the model, and visualizing the detected landmarks on the face, pose, and hands.

The core of the implementation is the LSTM-based neural network. The model is designed to learn temporal patterns from the sequence of keypoint values. It consists of multiple LSTM layers followed by dense layers. The model is compiled with the Adam optimizer and categorical cross-entropy loss.\newline

The LSTM (Long Short-Term Memory) layer is a type of recurrent neural network (RNN) layer that is capable of learning long-term dependencies. The equations governing an LSTM layer are as follows:

\begin{equation}
i_t = \sigma(W_i \cdot [h_{t-1}, x_t] + b_i)
\end{equation}
Input gate (\(i_t\)): Determines the amount of new information to be added to the cell state.

\begin{equation}
f_t = \sigma(W_f \cdot [h_{t-1}, x_t] + b_f)
\end{equation}
Forget gate (\(f_t\)): Decides what information from the cell state to discard.

\begin{equation}
o_t = \sigma(W_o \cdot [h_{t-1}, x_t] + b_o)
\end{equation}
Output gate (\(o_t\)): Controls the output from the LSTM cell.

\begin{equation}
C_t = f_t \ast C_{t-1} + i_t \ast \tanh(W_C \cdot [h_{t-1}, x_t] + b_C)
\end{equation}
Cell state (\(C_t\)): Updated cell state incorporating the input gate and forget gate decisions.

\begin{equation}
h_t = o_t \ast \tanh(C_t) 
\end{equation}
Hidden state (\(h_t\)): The output of the LSTM cell.\newline

The dense layer at the end of the network uses the softmax activation function to produce the final classification output:

\begin{equation}
y = \text{softmax}(W \cdot h + b)
\end{equation}

Where:
\begin{itemize}
    \item $\sigma$ is the sigmoid function
    \item $\tanh$ is the hyperbolic tangent function
    \item $\ast$ denotes element-wise multiplication
    \item $W$ and $b$ are weights and biases, respectively
    \item $x_t$ is the input at time step $t$
    \item $h_{t-1}$ is the hidden state from the previous time step
    \item $C_{t-1}$ is the cell state from the previous time step
\end{itemize}

The training process involves iterating over the dataset for 150 epochs. An early stopping callback is implemented to monitor categorical accuracy and halt training once a predefined threshold is reached, preventing the model from overfitting the training data.

The trained model is evaluated using a test set comprising 25\% of the data. The evaluation includes calculating a confusion matrix and accuracy score, providing insights into the model's performance on unseen data. The pose landmarker model in MediaPipe tracks 33 body landmark locations, as shown approximately in Figure 2. However, 11 of these landmarks are on the head, which is not relevant to our scenarios. Consequently, we excluded them from our model.

The final part demonstrates the real-time application of the trained model. The code captures live video frames, performs pose estimation, and feeds the sequence of keypoints into the LSTM model for action prediction. The predicted actions are displayed on the screen, along with a visual representation of the model's confidence in its predictions. An example of this is shown in Figure 3.

\begin{figure}[h]
\centering
\includegraphics[width=\linewidth]{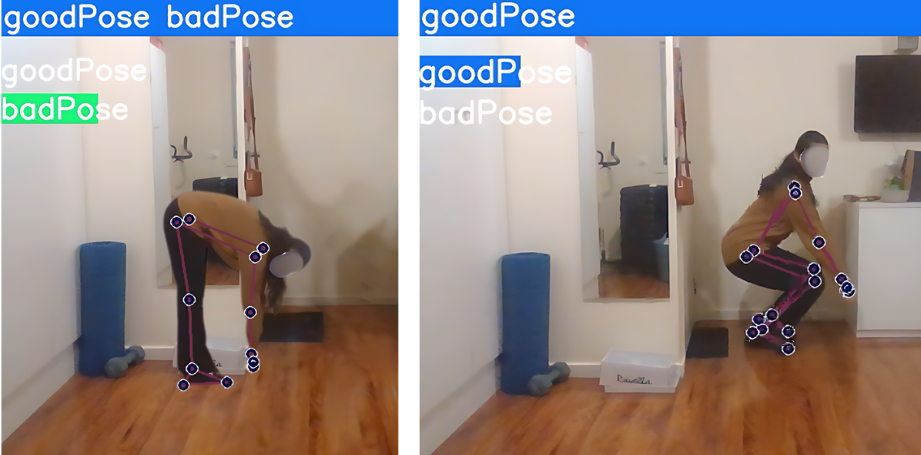}
\caption{Testing the Mediapipe model for each activity. a Bad Posture, b Good Posture.}
\end{figure}

\section{Results and Comparisons}
\label{sec:introduction}
\subsection{Visualizations}

The summary of the experiment results is provided in Table 2, which highlights key performance indicators for the proposed model. Notably, the model misclassified only one out of the 16 samples, achieving an impressive accuracy of 0.9565. Figure 4 illustrates the model's accuracy improvement over the training period, showcasing how accuracy evolves across epochs and providing insights into the learning trajectory. Figure 5 presents the Confusion Matrix, which details true positives, true negatives, false positives, and false negatives, offering a comprehensive view of the model's classification accuracy and error distribution. Additionally, Figure 6 displays the ROC curve, highlighting the trade-off between sensitivity and specificity for the classification model.

\begin{table}
\centering
\caption{Model Training and Evaluation Metrics}
\label{tab:model_metrics}
\begin{tabular}{ccc}
\toprule
Metric & Value & Comments \\
\midrule
Epochs & 150 & No. of training epochs \\
Categorical Accuracy & 0.9565 & Accuracy on training data \\
False Positives & 1 & No. of false positive predictions \\
True Negatives & 1 & No. of true negative predictions \\
True Positives & 1 & No. of true positive predictions \\
Final Accuracy Score & 0.9375 & Accuracy on evaluation dataset \\
\bottomrule
\end{tabular}
\end{table}

\begin{figure}[h]
\centering
\includegraphics[width=\linewidth]{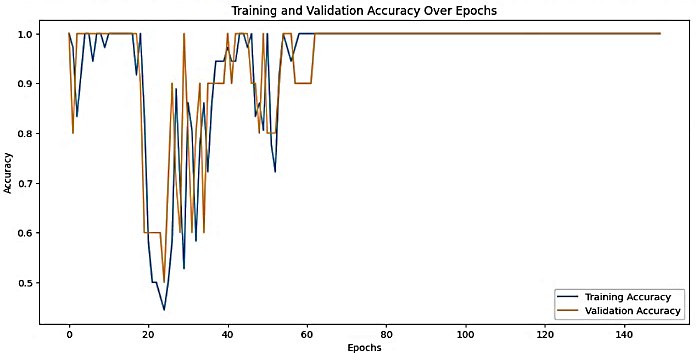}
\caption{Performance visualization of the Training and Validation Accuracy over Epochs.}
\end{figure}

\begin{figure}[h]
\centering
\includegraphics[width=\linewidth]{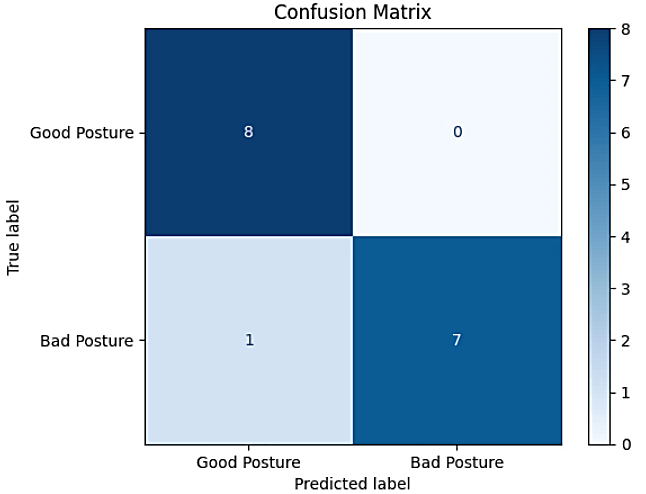}
\caption{Performance visualization of the Training and Validation Accuracy over Epochs.}
\end{figure}
 
\begin{figure}[h]
\centering
\includegraphics[width=\linewidth]{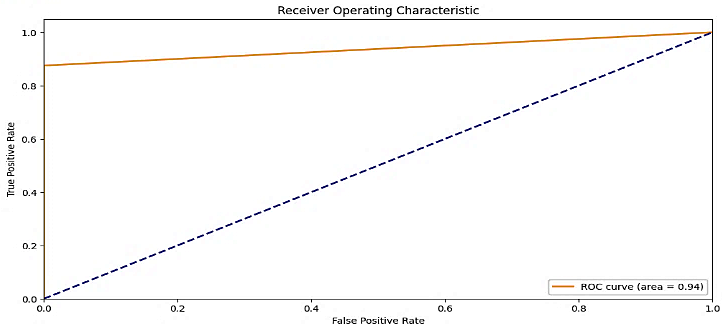}
\caption{Performance visualization of the ROC Graph.}
\end{figure}

\subsection{Comparative Analysis of Posture Detection Models}
In the realm of real-time posture monitoring systems utilizing LSTM-based neural networks, it is imperative to conduct a thorough comparison with existing solutions to establish the uniqueness and efficacy of the proposed model. This comparative analysis delves into a meticulous examination of two pertinent works: "Tennis Shot Identification Using YOLOv7 Pose Estimation and LSTM" and "LSTM Pose Machines." \cite{shu2022soft} The evaluation focuses on the methodologies employed, the performance achieved, and the applicability of these models within the domain of posture detection.

\subsubsection{Existing Work 1: Tennis Shot Identification Using YOLOv7 Pose Estimation and LSTM}
The methodology employed in the referenced work involves several critical components for pose estimation and sequence modeling, specifically designed for tennis shot identification. The approach begins with YOLOv7 for detecting body keypoints, forming the foundation for accurately capturing the nuances of various tennis shots.

An LSTM model is then utilized to process the sequence of detected keypoints, allowing for a comprehensive analysis of the player's movements throughout the shot execution process. The model's architecture is noteworthy, featuring 3 ConvLSTM2D layers and 3 MaxPool3D layers, which enhance its ability to recognize the intricate movement patterns characteristic of tennis shots. The dataset used in this methodology consists of meticulously annotated tennis shot videos, providing a robust basis for training and validation.

The methodology demonstrates commendable accuracy in identifying different types of tennis shots, highlighting the effectiveness of the pose detection approach using YOLOv7. The robustness of the pose detection technique is particularly evident in its ability to capture the dynamic and high-speed movements inherent in sports activities, underscoring its applicability in such scenarios.

However, while the methodology excels in tennis shot identification, certain limitations must be acknowledged. The domain-specific nature of the approach limits its generalizability to other activities or postures outside of tennis. Additionally, the model's performance is closely tied to the specificity of the dataset used, which may hinder its adaptability to diverse datasets or scenarios. Another significant limitation is the absence of real-time feedback mechanisms, a critical feature for applications such as manual lifting tasks. Real-time feedback and risk categorization are essential for these tasks, which are not directly addressed by the discussed methodology.

\begin{table*}
\centering
\caption{Comparative experiments with different models.}
\label{tab:comparative_experiments}
\begin{tabular}{ccc}
\toprule
Model & Categorical Accuracy & Accuracy Score \\
\midrule
Proposed LSTM-Based Model & 0.9565 & 0.9375 \\
Tennis Shot Identification (YOLOv7 + LSTM) & 0.8615 & 0.8536 \\
Yoga LSTM Pose Machines & 0.8348 & 0.8085 \\
\bottomrule
\end{tabular}
\end{table*}

\subsubsection{Existing Work 2: Yoga LSTM Pose Machines}
The methodology outlined in the described work integrates key elements of pose estimation and sequence modeling, with a particular focus on diverse human activities annotated with keypoints. The pose estimation component utilizes OpenPose for robust keypoint detection, providing the foundation for accurate posture analysis. This is followed by sequence modeling, which combines multiple CNN and LSTM layers to process temporal sequences of keypoints, enabling a comprehensive analysis of human movements across various activities. The dataset supporting this methodology comprises a wide range of human activities, meticulously annotated with keypoints using image and frame data, ensuring diverse training and validation scenarios.

The methodology demonstrates commendable accuracy, particularly in recognizing various yoga poses, showcasing its effectiveness in modeling temporal dependencies through multiple LSTM layers. Its ability to capture the subtleties of human movement during yoga practice underscores its potential in action recognition tasks.

However, the methodology also presents certain limitations that merit attention. When tested in real-time scenarios, the approach showed inaccuracies, especially in handling the continuous and varied nature of yoga postures, posing challenges for real-time applications. The integration of multiple LSTM layers contributes to increased computational complexity, necessitating robust computational resources for efficient execution. Additionally, the generalizability of the model across different datasets and activity types remains a concern, as variations in data characteristics and activity contexts may significantly impact performance outcomes.

\subsubsection{Comparative Analysis}

Our proposed model features a sophisticated architecture consisting of three LSTM layers and three dense layers, specifically designed to capture temporal dependencies and ensure precise posture classification. This system seamlessly integrates a computer vision component for continuous posture data capture, an LSTM-based model for sequential data analysis, and a user-friendly web interface that provides real-time feedback to users. A key innovation of our solution is the incorporation of a risk analysis framework, which categorizes postures based on risk levels and delivers immediate feedback on lifting techniques.

Compared to existing approaches such as YOLOv7 with LSTM and LSTM Pose Machines, our model offers distinct advantages. It is purpose-built for handling continuous and varied postures in real-time, a capability that is not fully addressed by other solutions. Additionally, the integration of a risk analysis framework enhances the model’s practicality, a feature often lacking in current models. This combination of improved accuracy, real-time responsiveness, and risk assessment positions our model as a pioneering solution in posture monitoring systems.

The benefits of deploying our model are extensive. Primarily, it delivers superior accuracy in real-time posture correction, significantly reducing the risk of musculoskeletal disorders (MSDs). Enhanced temporal data analysis ensures more precise posture tracking and correction, contributing to a safer work environment. This, in turn, lowers healthcare costs related to MSDs and boosts worker productivity. The inclusion of real-time feedback mechanisms and a user-friendly interface further enhances the user experience, promoting adherence to safe lifting practices and fostering a culture of workplace safety.

Comparative analysis reveals compelling results, highlighting our model's performance in real-time posture monitoring. The proposed LSTM-based model emerged as the top performer, achieving a categorical accuracy of 0.9565 and an accuracy score of 0.9375. In comparison, the Tennis Shot Identification model using YOLOv7 and LSTM achieved a categorical accuracy of 0.8615 and an accuracy score of 0.8536, while LSTM Pose Machines scored 0.8348 in categorical accuracy and 0.8085 in accuracy score.

As shown in Table 3, the results underscore the superiority of our LSTM-based model in terms of accuracy and generalizability. This success is attributed to several factors, including the use of a diverse dataset, iterative data collection processes, and the model's streamlined architecture. While existing models perform well within their specialized domains, they face challenges related to generalizability and computational complexity. In contrast, our model excels in simplicity, efficiency, and accuracy, making it a highly promising solution for real-time posture monitoring across various contexts.

These comparative experiments validate the effectiveness of our proposed method and establish its significance in posture monitoring and ergonomic assessments. The model's ability to outperform existing solutions underscores its importance as a practical and efficient tool for enhancing workplace safety and mitigating the risk of musculoskeletal disorders.

\section{Conclusion and Future Work}
In conclusion, this research offers valuable insights and practical solutions for improving occupational health and safety, particularly in the context of manual lifting tasks. The findings highlight the critical role of advanced technologies in posture monitoring and risk assessment, suggesting a promising approach to reducing the risk of musculoskeletal disorders (MSDs) and enhancing workplace safety.

Future work will focus on addressing the identified limitations, expanding the dataset, and refining the system for broader deployment. Key areas for further exploration include improving data collection and annotation processes to enhance model accuracy and adaptability across various environments. Additionally, optimizing model performance while balancing complexity with real-time capabilities will be crucial. Enhancing keypoint calculations and leveraging advanced machine learning and computer vision techniques will be essential to increase the model's robustness.

Improving the user experience through refined interfaces and feedback mechanisms will also be a priority. Streamlining these interfaces and incorporating real-time feedback will enhance both usability and effectiveness. Future integration with Internet-of-Things (IoT) devices offers exciting opportunities for real-time posture monitoring and personalized feedback, potentially increasing user engagement and adherence to recommended postures.

Long-term efficacy studies will be conducted in collaboration with healthcare experts to evaluate the sustained impact of the posture correction system on health and safety. Additionally, efforts will be made to scale and deploy the system across diverse environments to ensure widespread adoption and impact.

\bibliographystyle{ACM-Reference-Format}
\bibliography{sample-base}

\end{document}